\documentclass[10pt,onecolumn]{IEEEtran}

\usepackage{amsmath}
\usepackage{array}
\usepackage{booktabs}
\usepackage{cite}
\usepackage{graphicx}
\usepackage{siunitx}
\usepackage{url}
\usepackage{xcolor}
\usepackage{threeparttable}
\usepackage{placeins}
\usepackage[hidelinks]{hyperref}

\graphicspath{{figures/}}

\title{Underwater Dead Reckoning with Deployable Situation-Triggered Covariance Scheduling}

\hypersetup{
  pdftitle={Underwater Dead Reckoning with Deployable Situation-Triggered Covariance Scheduling},
  pdfauthor={Akshay Naik, Ramavarapu S. Sreenivas, Dustin Nottage, and Ahmet Soylemezoglu},
  pdfkeywords={dead reckoning, Doppler velocity log, situation-aware estimation, underwater navigation}
}

\author{Akshay~Naik,
Ramavarapu~S.~Sreenivas,
Dustin~Nottage,
and~Ahmet~Soylemezoglu%
\thanks{A. Naik is with the Dept. of Electrical and Computer Engineering, University of Illinois Urbana-Champaign, 306 N. Wright St., Urbana, IL 61801 USA (e-mail: akshayn3@illinois.edu). R. S. Sreenivas is with the Dept. of Industrial and Enterprise Systems Engineering, University of Illinois Urbana-Champaign, 117 Transportation Building, 104 S. Mathews Ave., Urbana, IL 61801-3080 USA. D. Nottage and A. Soylemezoglu are with the U.S. Army Engineer Research and Development Center, Construction Engineering Research Laboratory, 2902 Newmark Dr., P.O. Box 9005, Champaign, IL 61826-9005 USA.}%
}

\begin{document}
\maketitle

\begin{abstract}
Underwater dead reckoning estimates vehicle position when vision is unavailable and external positioning cannot be assumed. A single set of filter parameters can work well in many situations, but fixed tuning may be poorly matched during turns, motion transitions, or periods when sensor measurements are less reliable. This paper presents the Situation-Triggered Calibrated Adaptive Robust Extended Kalman Filter for a BlueROV2. An onboard probabilistic trigger identifies the current motion situation while one error-state filter runs continuously. When the trigger is confident, the filter changes only to the corresponding pre-calibrated process- and measurement-noise matrices; the state estimate, covariance history, dynamics, and measurement models are not reset or replaced. The trigger, noise profiles, and a one-time Doppler velocity log yaw-alignment correction are calibrated offline using sparse AprilTag-supervised pool runs. A separate validation set selects the scheduling policy, which is then fixed before held-out testing. Across four held-out pool runs, the method reduces label-weighted mean per-run translation root-mean-square error from \SI{0.488}{\meter} to \SI{0.471}{\meter} relative to the same filter backbone with one global noise profile, and every held-out run favors the scheduled method. A paired bootstrap over 10-second segments gives a candidate-minus-baseline difference of \SI{-0.017}{\meter} with a 95\% confidence interval of $[-0.024,-0.008]~\si{\meter}$, while orientation error remains essentially unchanged. These results indicate that situation-aware covariance scheduling provides a modest but consistent vision-free dead-reckoning improvement without switching estimators or resetting the filter.
\end{abstract}


\section{Introduction}
Underwater dead reckoning (DR) is constrained by the absence of the measurements that many robotic systems use as anchors. Satellite navigation is unavailable below the surface. Motion-capture systems remain confined to instrumented facilities. Acoustic positioning can provide global information, but it introduces infrastructure, calibration, cost, and failure modes of its own \cite{Kinsey2006Survey,Paull2014AUVReview}. A low-cost remotely operated vehicle must therefore rely on onboard signals such as inertial measurements, Doppler velocity log (DVL) velocity when bottom lock is healthy, depth and altitude, and commanded motion. These signals support short-term localization, but their integrated errors grow over time. The central question is how much drift can be reduced using only the sensors available at runtime.

That drift does not accumulate uniformly. Hovering near the bottom, executing a turn, and translating while DVL returns are stale or invalid induce different uncertainty patterns. A covariance model suited to clean straight motion can be overconfident during turns, whereas a model that remains conservative through turns can underuse reliable measurements during straight segments. This is the motivation for situation awareness in dead reckoning: rather than use one static description of sensor trust, the estimator can condition its fusion behavior on the current motion regime.

Existing situation-aware estimators commonly follow one of two deployment paths. Some switch between separate estimators or run banks of submodels \cite{BlomBarShalom1988IMM,Li2016MultiModelEKF,Zhang2022IIMMUKF}, which increases runtime complexity and can make certification or failure analysis more difficult. Others adapt continuously but depend on quantities, such as estimated error or innovation statistics, that may be difficult to validate under sparse supervision \cite{Kang2026RMKAKF}. This paper takes a narrower path. It keeps one continuous error-state extended Kalman filter (EKF) and changes only its \emph{fixed matrices}: the process noise $Q$ and the measurement noise $R$. The propagated state $\mathbf{x}$ and covariance $\mathbf{P}$ persist across every switch, so the filter does not reset its estimate; it only changes the uncertainty assigned to subsequent propagation and measurement updates. The globally tuned backbone is the Calibrated Adaptive Robust Extended Kalman Filter (CAR-EKF), and the situation-triggered variant is the Situation-Triggered Calibrated Adaptive Robust Extended Kalman Filter (ST-CAR-EKF). Its probabilistic trigger is trained offline against ground-truth motion labels but reads only onboard sensors at runtime, behind a confidence gate chosen on validation.

A method that calibrates noise on ground truth raises an important question: how much should a calibration fitted on a limited set of supervised runs be expected to transfer? We therefore frame the calibration explicitly as a small transfer-learning problem in the sense of carrying information from source runs to held-out target runs \cite{PanYang2010TransferLearning}. The noise matrices and a one-time DVL misalignment correction are estimated on a small set of calibration runs and then frozen; the experiment is whether that frozen description transfers to held-out runs from the same pool benchmark. This framing makes the generalization boundary explicit and turns the amount of required supervision into an empirical quantity rather than an assumption.

Obtaining that supervision underwater is also nontrivial, because reliable pose labels are difficult to acquire below the surface. We drive a BlueROV2 in a swimming pool over a small set of unsurveyed floor AprilTags \cite{Olson2011AprilTag,Krogius2019FlexibleTags,Bauschmann2023UnderwaterAprilTag}. The tags are not the contribution. They are an offline measurement instrument that produces sparse six-degree-of-freedom pose labels after a run, and they are used only for calibration, situation labeling, and scoring. At deployment the filter is strictly vision-free: no camera, no tag detections, no map, and no ground truth enter the runtime path. Everything the estimator needs to localize comes from the inertial and DVL streams that the vehicle already produces.

This paper makes four contributions.
\begin{itemize}
\item \textbf{ST-CAR-EKF.} A runtime covariance-scheduling estimator that keeps one continuous error-state EKF and swaps only fixed $Q$/$R$ matrices when an onboard trigger is confident. State and covariance are never reset at a switch.
\item \textbf{Sparse offline calibration.} We use AprilTag ground truth offline to fit a one-time DVL alignment, calibrate fixed noise profiles, and label situations; the frozen runtime estimator remains vision-free.
\item \textbf{Validation-only trigger policy.} A probabilistic situation trigger runs on onboard features only, and per-situation activation gates are selected on validation rather than tuned on held-out test.
\item \textbf{Reproducible dead-reckoning benchmark.} We report label-weighted absolute pose error, per-run held-out deltas, paired bootstrap uncertainty, fixed-horizon relative-pose drift, drift per distance traveled, and tail errors, with the commercial stream restricted to black-box common-row comparisons.
\end{itemize}

\section{Related Work}
\subsection{Underwater Navigation as a Sensor-Fusion Problem}
Underwater navigation is organized around a persistent constraint: position is difficult to observe directly below the surface. Kinsey, Eustice, and Whitcomb framed the field as a blend of sensor technology, deterministic navigation, and stochastic estimation built from inertial sensing, Doppler sonar, pressure, magnetic heading, and acoustic positioning \cite{Kinsey2006Survey}. Paull et al.\ later cast autonomous underwater vehicle (AUV) localization as a set of tradeoffs among dead reckoning, acoustic localization, geophysical navigation, and simultaneous localization and mapping (SLAM), with DVL-aided inertial navigation as the usual short-term backbone \cite{Paull2014AUVReview}. Recent IEEE Journal of Oceanic Engineering work keeps that systems perspective, evaluating multisensor fusion strategies and navigation aided by terrain, bathymetry, and acoustic baseline structure \cite{Bucci2023UKFFusion,Salavasidis2021TerrainAided,Hong2025BathymetryBayesian,Wang2025HybridBaseline}. A reproducible BlueROV2 pool dataset with raw DVL, inertial measurement unit (IMU), range, depth, commands, and sparse pose labels sits in a small but underserved corner of that map: low cost, vision-free at runtime, and explicit about its limited supervision.

\subsection{Robust and Adaptive Inertial/DVL Filtering}
DVL returns underwater are not the clean Gaussian observations textbooks assume. Bottom lock degrades, individual beams drop out, mounting and scale errors creep in, and the effective noise depends on water and motion \cite{CohenKlein2024SeamlessDVL}. Robust filtering traces back to residual down-weighting in the Huber tradition \cite{Huber1964Robust}, while adaptive Kalman methods adjust noise covariances from innovations, residuals, or approximate Bayesian posteriors \cite{SarkkaHartikainen2013VB,Akhlaghi2017AdaptiveNoise}. Underwater inertial navigation system (INS)/DVL work follows the same logic: Kang et al.\ combine Sage--Husa adaptation with Huber residual weighting \cite{Kang2026RMKAKF}, and learned process-noise regressors such as ProNet adapt INS/DVL covariance from inertial features \cite{OrKlein2022ProNet}. Our backbone draws on this family, but our claim is orthogonal to designing a heavier single filter. We ask whether \emph{discretely retargeting} a strong calibrated filter's noise model to the current situation, triggered from onboard features, can recover accuracy that a validation-selected static configuration leaves on the table.

\subsection{Factor-Graph Smoothing and Batch Estimation}
Factor graphs express navigation when history should be optimized jointly rather than folded forward one update at a time, building on iSAM2 and IMU preintegration \cite{Kaess2012ISAM2,Forster2015Preintegration} and extending to INS, ultra-short baseline (USBL), and DVL fusion \cite{Li2023RobustGraph,Qin2023FGO,Song2023FGOILNS}, DVL-INS with laser loop closures \cite{AlBaali2023DVLINSLoopClosure}, and adaptive non-Gaussian underwater noise \cite{Cheng2026ACEFGO}. Smoothing is an excellent offline yardstick, but a deployed estimator may not be permitted to revise the past at will, so we treat batch smoothing as a reference rather than as the live method.

\subsection{Learning for Inertial and DVL-Aided Odometry}
Learning has sharpened inertial odometry \cite{Herath2020RoNIN,Liu2020TLIO,Qiu2023AirIMU}, DVL velocity estimation and missing-beam handling \cite{Cohen2022BeamsNet,Yona2024MissBeamNet,CohenKlein2024SeamlessDVL}, and adaptive INS/DVL covariance selection \cite{OrKlein2022ProNet}, and recent work is careful about how learned outputs interact with filter consistency \cite{Cohen2025CrossCorrelation}. These methods suggest that learning is most defensible when it corrects or selects within a model-based estimator rather than replacing it. ST-CAR-EKF follows that structure: learning enters only as a situation trigger that selects a noise model, while the dead-reckoning backbone stays a causal, inspectable filter.

\subsection{Fiducial Markers and Underwater Ground Truth}
AprilTags give repeatable six-degree-of-freedom pose from a calibrated image \cite{Olson2011AprilTag,Krogius2019FlexibleTags} and slot naturally into graph SLAM \cite{Pfrommer2019TagSLAM}. Underwater they are less forgiving: accuracy depends on calibration medium, range, and motion \cite{Bauschmann2023UnderwaterAprilTag}, and public underwater datasets with trustworthy six-degree-of-freedom ground truth remain rare \cite{Xu2026TankDataset}. Our use is deliberately narrow. A few unsurveyed pool-floor tags provide the offline reference labels; they never enter the runtime estimator.

\subsection{Multiple-Model and Situation-Aware Navigation}
The observation that one model may not fit all motion regimes is well established: interacting multiple-model filtering \cite{BlomBarShalom1988IMM}, multi-model EKF \cite{Li2016MultiModelEKF}, and interacting multiple-model unscented Kalman filter (UKF) methods \cite{Zhang2022IIMMUKF} run banks of submodels with probabilistic association. ST-CAR-EKF differs in three ways that matter for deployment. It runs one filter rather than a bank, it changes only the fixed noise matrices rather than the state model, and it triggers from onboard features under a confidence gate rather than from errors it could not observe at sea.

\section{Dataset and Ground-Truth Construction}
\subsection{Collection Setting}
The BlueROV2 has a manufacturer-specified nominal body length of \SI{0.457}{\meter}~\cite{BlueRoboticsBlueROV2Datasheet}, which is used below only as a physical scale reference. The primary dataset comprises BlueROV2 runs collected in a swimming pool with four AprilTags resting on the floor. Each Robot Operating System (ROS) bag records a raw camera stream and the onboard topics that include IMU, DVL twist and quality fields, altitude, range, static pressure, and teleoperation commands. A later group of runs additionally records a commercial DVL dead-reckoning stream: a logged onboard output whose internal model is unavailable. We therefore treat this commercial stream as a black-box reference, not a reproducible baseline, and score it only on available common rows.

\subsection{Runtime Data Boundary}
The runtime boundary is intentionally narrow. At deployment, the estimator reads only onboard ROS streams: inertial messages, attitude and heading reference system (AHRS) messages, DVL velocity and quality fields, altitude, range, pressure, and teleoperation commands. A causal backward as-of merge converts those streams into \SI{10}{\hertz} features. Each sample carries age and freshness metadata, and samples older than \SI{0.75}{\second} are rejected.

The EKF uses IMU and control cues for propagation plus DVL and altitude/range updates. The trigger uses onboard features derived from angular rate, DVL velocity, figure of merit (FOM), validity, altitude or range, commanded motion, and sample freshness. The frozen primary trigger uses 31 selected onboard features, dominated by angular-rate structure and DVL velocity/FOM terms. AprilTag detections, the recovered tag map, camera poses, ground-truth residuals, and the commercial stream are excluded from runtime estimation; they enter only as offline calibration labels, scoring labels, or black-box comparison artifacts. A topic-level contract is included in the supplemental artifacts.

\subsection{Tag Map Recovery and Pose Labels}
\begin{figure}[!htbp]
\centering
\includegraphics[width=\textwidth]{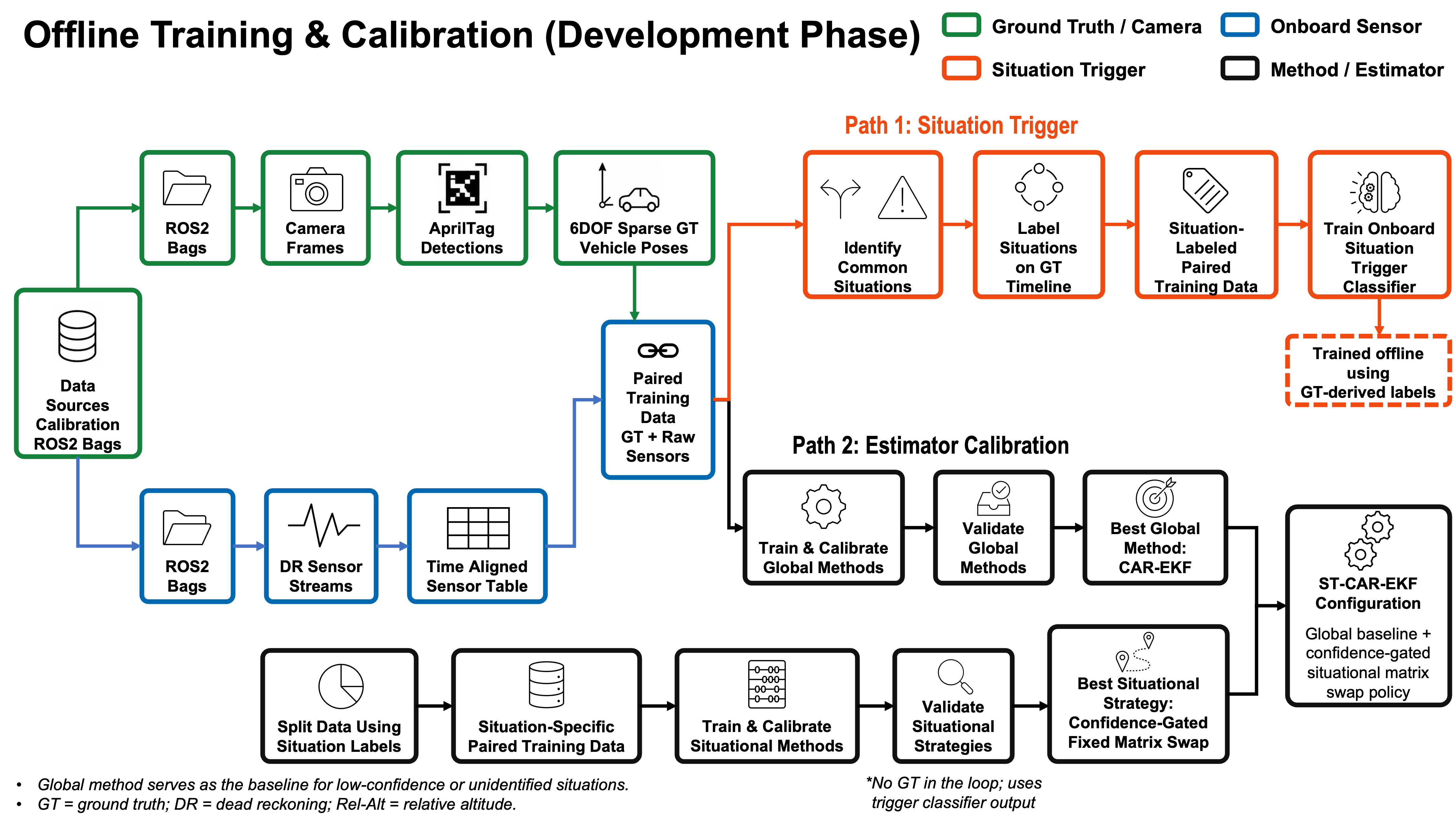}
\caption{Data processing and model training pipeline (offline only). \textbf{Green:} sparse AprilTag ground-truth poses from camera frames. \textbf{Blue:} time-aligned IMU/DVL sensor streams paired with labels. \textbf{Orange (Path~1):} situation identification, GT timeline labeling, and onboard trigger-classifier training---supervised with labels offline but deployable from onboard features at runtime. \textbf{Black:} global method selection (CAR-EKF), situation-specific fixed-matrix calibration, and validation, yielding the ST-CAR-EKF configuration. AprilTag labels score trajectories post hoc only; they never enter the runtime path.}
\label{fig:pipeline}
\end{figure}

\begin{figure}[!htbp]
\centering
\includegraphics[width=0.66\linewidth,height=0.44\textheight,keepaspectratio]{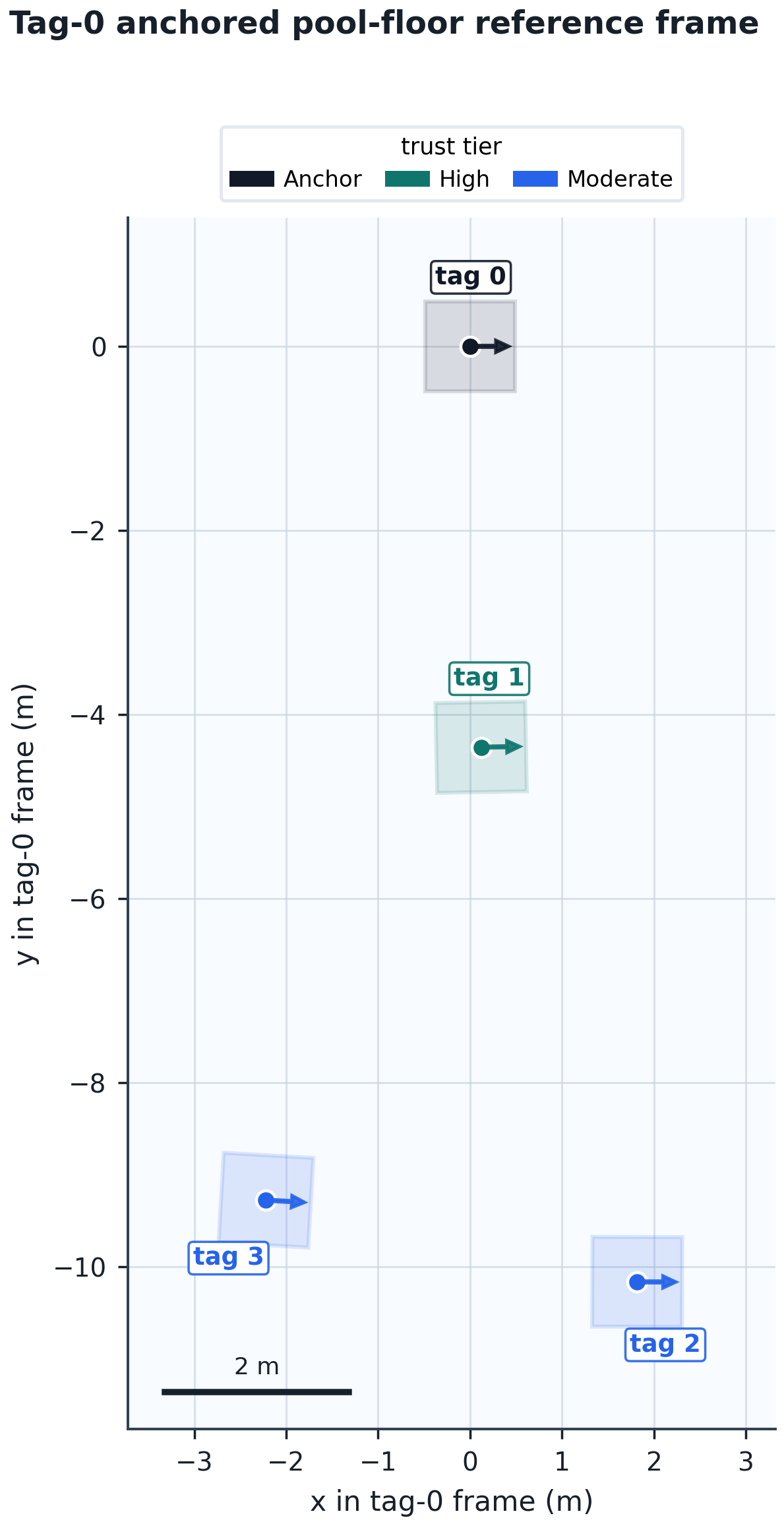}
\caption{Trusted AprilTag map used to generate sparse camera and vehicle pose labels. Tag~0 defines the origin and anchors the reference frame; each square shows physical tag extent and yaw, colored by trust tier.}
\label{fig:tag-layout}
\end{figure}

The tag map and resulting ground-truth (GT) pose labels are reconstructed offline. Fig.~\ref{fig:pipeline} summarizes the full offline training and calibration flow: AprilTag pose labels and time-aligned sensor streams are paired, situations are labeled from ground-truth kinematics, global and situational noise models are calibrated and validated, and the frozen ST-CAR-EKF policy (global CAR-EKF baseline plus confidence-gated situational matrix swap) is selected before any held-out evaluation. Tag~0 defines the world origin, and co-visibility chains propagate the remaining tags into the same frame, retaining full six-degree-of-freedom pose because the pool floor and tag mounts are not assumed planar (Fig.~\ref{fig:tag-layout}). For each usable camera frame, detections become camera poses in the tag frame and then base-link poses through fixed extrinsics. The aligned dataset pairs each labeled timestamp with the most recent fresh sensor sample inside a staleness bound, and a separate \SI{10}{\hertz} fixed-rate table supports runtime-style feature construction.

\subsection{Curation and Splitting}
\begin{figure}[!htbp]
\centering
\includegraphics[width=0.86\textwidth]{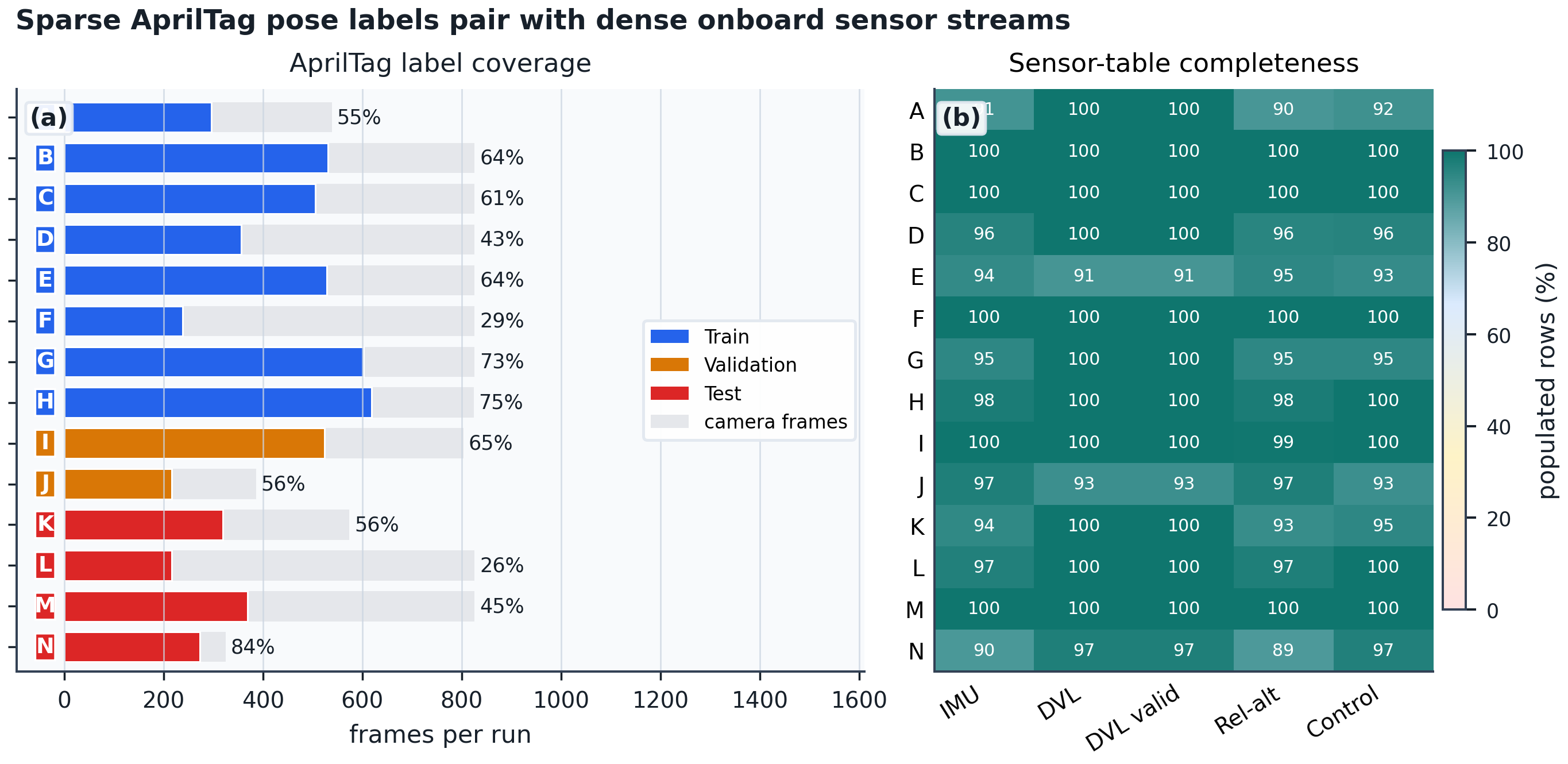}
\caption{Curated combined dataset. (a) AprilTag-derived pose labels are a sparse fraction of each camera stream; run letters and bar color encode the train/validation/test split. (b) Completeness of the aligned onboard sensor groups the deployable estimator consumes at runtime.}
\label{fig:dataset-coverage}
\end{figure}

Splits are made at the bag level so that no frame leaks between training and evaluation. The curated combined dataset holds 14 bags (8 train, 2 validation, 4 test), 10{,}059 camera-frame rows, and 5{,}595 labeled pose rows, with mean onboard-sensor feature coverage near \SI{97}{\percent} (Table~\ref{tab:dataset}, Fig.~\ref{fig:dataset-coverage}). The split is motion-stratified so the train and test situation distributions stay close. Only three trusted runs carry the commercial stream, so black-box reference comparisons are reported on those matched runs rather than as broad statistical claims.

\begin{table}[!t]
\centering
\caption{Curated Combined Dataset Summary}
\label{tab:dataset}
\begin{tabular}{lrrr}
\toprule
Split & Bags & Camera rows & Label rows\\
\midrule
Train & 8 & 6319 & 3677\\
Validation & 2 & 1189 & 740\\
Test & 4 & 2551 & 1178\\
\midrule
Total & 14 & 10059 & 5595\\
\bottomrule
\end{tabular}
\end{table}

\section{Situations and Onboard Triggers}
\subsection{Situation Labeling from Ground Truth}
We define a small set of motion situations that plausibly call for different noise models: \emph{straight line}, \emph{turning}, \emph{transition/mixed}, and \emph{hover/low speed}. Each labeled timestamp is assigned a primary situation from ground-truth kinematics, using smoothed speed, yaw rate, and vertical motion with a fixed priority order; timestamps without a confident pose are marked untrusted and dropped. Because these labels come from ground truth, they exist only offline, which is precisely why they cannot be the runtime signal. Fig.~\ref{fig:pipeline} (orange Path~1) shows that offline labeling and trigger-training chain.

\subsection{Probabilistic Triggering from Onboard Features}
\begin{figure}[!htbp]
\centering
\includegraphics[width=0.80\textwidth]{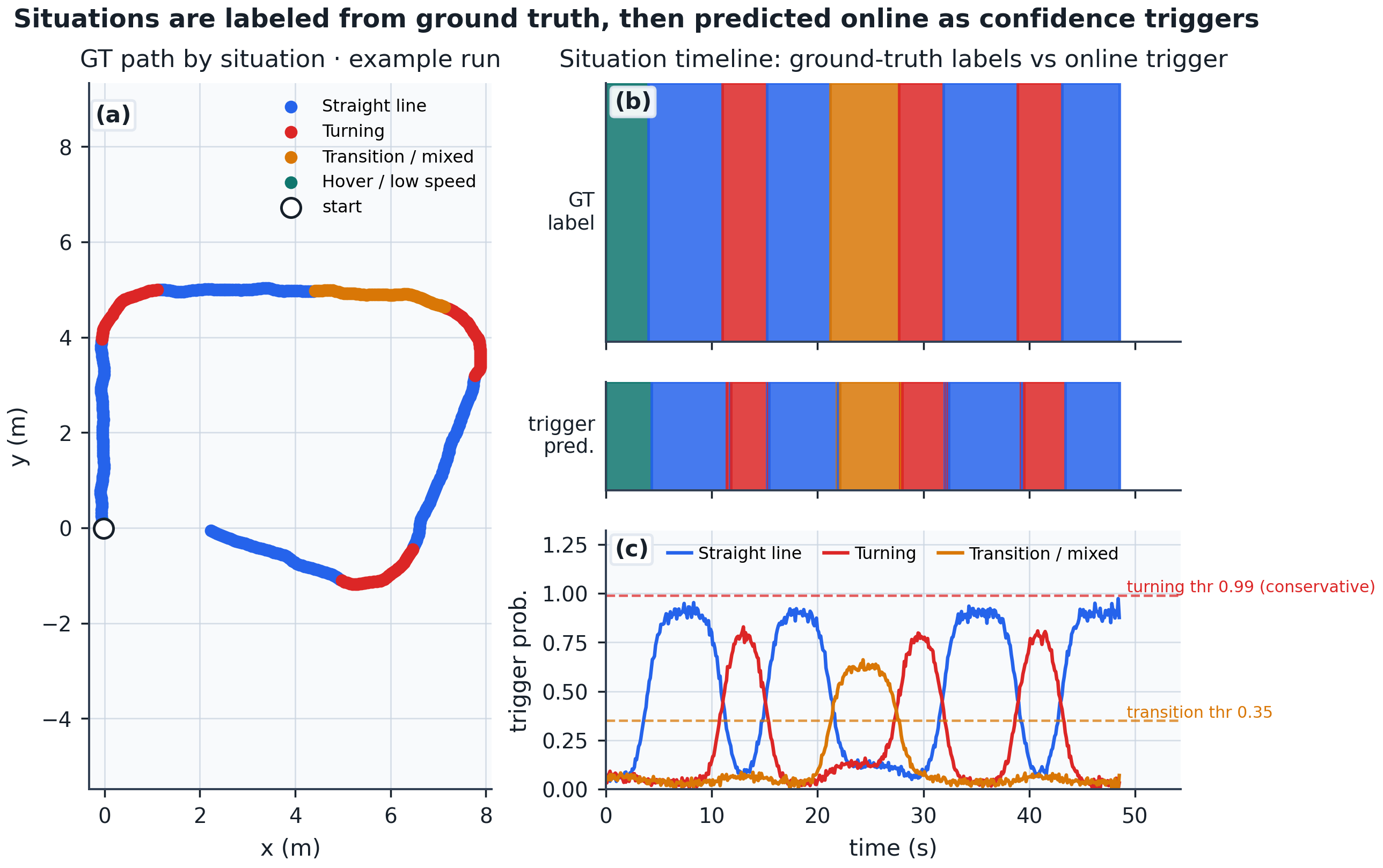}
\caption{Situation triggering on a clean illustrative run. (a) The path is segmented into motion situations from GT kinematics. (b) GT labels (top) supervise an onboard classifier whose predicted situation (bottom) is what is actually available at deployment, with no ground truth. (c) Per-situation trigger probabilities with the frozen deployable activation thresholds. Turning uses assertive-fusion but only when $P(\text{turning})\ge 0.99$; lower-confidence turning predictions keep the global profile until a lower gate is shown to transfer on held-out pool runs.}
\label{fig:situations}
\end{figure}

To make situations usable at runtime, we train a probabilistic classifier that consumes only onboard features, angular rate, DVL validity and velocity, altitude and range, commands, and sample freshness, and emits a per-step probability for each situation along with a max-probability confidence. The classifier is supervised by the ground-truth situation labels but never sees ground truth at inference. At deployment the trained trigger and frozen ST-CAR-EKF policy run on live sensor streams only. Fig.~\ref{fig:situations} walks through trigger behavior on a clean illustrative run: the path segmented by situation (a), the ground-truth label band against the online trigger prediction (b), and the per-situation trigger probabilities with their activation thresholds (c).

\FloatBarrier
\section{Method: ST-CAR-EKF}
Both the global baseline and our method are drawn from the same \emph{Calibrated Adaptive Robust EKF} (CAR-EKF) family: one continuous error-state filter whose search space includes fixed, adaptive, and robust covariance configurations. The validation-selected global member for this benchmark uses fixed covariances, while ST-CAR-EKF introduces adaptation through situation-triggered covariance scheduling. ST-CAR-EKF is therefore not a different filter family; it is a different scheduling policy on the same backbone.
\subsection{Calibrated EKF Backbone}
The backbone is an error-state EKF that propagates pose and velocity from IMU and control and updates from DVL velocity and altitude/range. Its configuration is drawn from a calibrated robust family: the search space includes the process-noise scalars, the measurement-noise scalars, Huber robust weighting of residuals \cite{Huber1964Robust}, and innovation-adaptive measurement noise \cite{Akhlaghi2017AdaptiveNoise}. Two calibration choices shape every result that follows. First, the DVL frame is corrected once by a yaw misalignment of about \SI{1.8}{\degree}, estimated in closed form from 64 sparse velocity pairs across the eight training runs; the fit is yaw-only, with unit scale. Second, the noise configuration is chosen by minimizing validation translation root-mean-square error (RMSE) over the candidate family. On this comparatively benign pool data that search selected calibrated \emph{fixed} noise and left the Huber and adaptive-noise options switched off, an empirical outcome we return to in Section~\ref{sec:discussion} because it indicates which covariance choices are supported by this benchmark.

\subsection{Situation Fixed-Matrix Profiles}
For each situation we calibrate an alternative set of fixed matrices, a \emph{profile}, that is best for that situation in isolation. The two profiles used by the deployable policy have direct physical interpretations. The \emph{calm-process} profile lowers the velocity random-walk and gyro process noise (for example, velocity random walk from \SI{0.24}{} to \SI{0.18}{\meter\per\second\squared}), increasing reliance on propagation during clean straight cruising. The \emph{assertive-fusion} profile instead tightens the measurement noise (for example, DVL planar noise from \SI{0.045}{} to \SI{0.030}{\meter\per\second}), giving greater weight to DVL updates through transitions and turns. Crucially, a profile changes only $Q$ and $R$; the state-transition and measurement models are identical across profiles. That invariance is what lets the filter change profiles mid-run without any reset.

\subsection{Confidence-Gated Switching}
\begin{figure}[!htbp]
\centering
\includegraphics[width=0.94\textwidth]{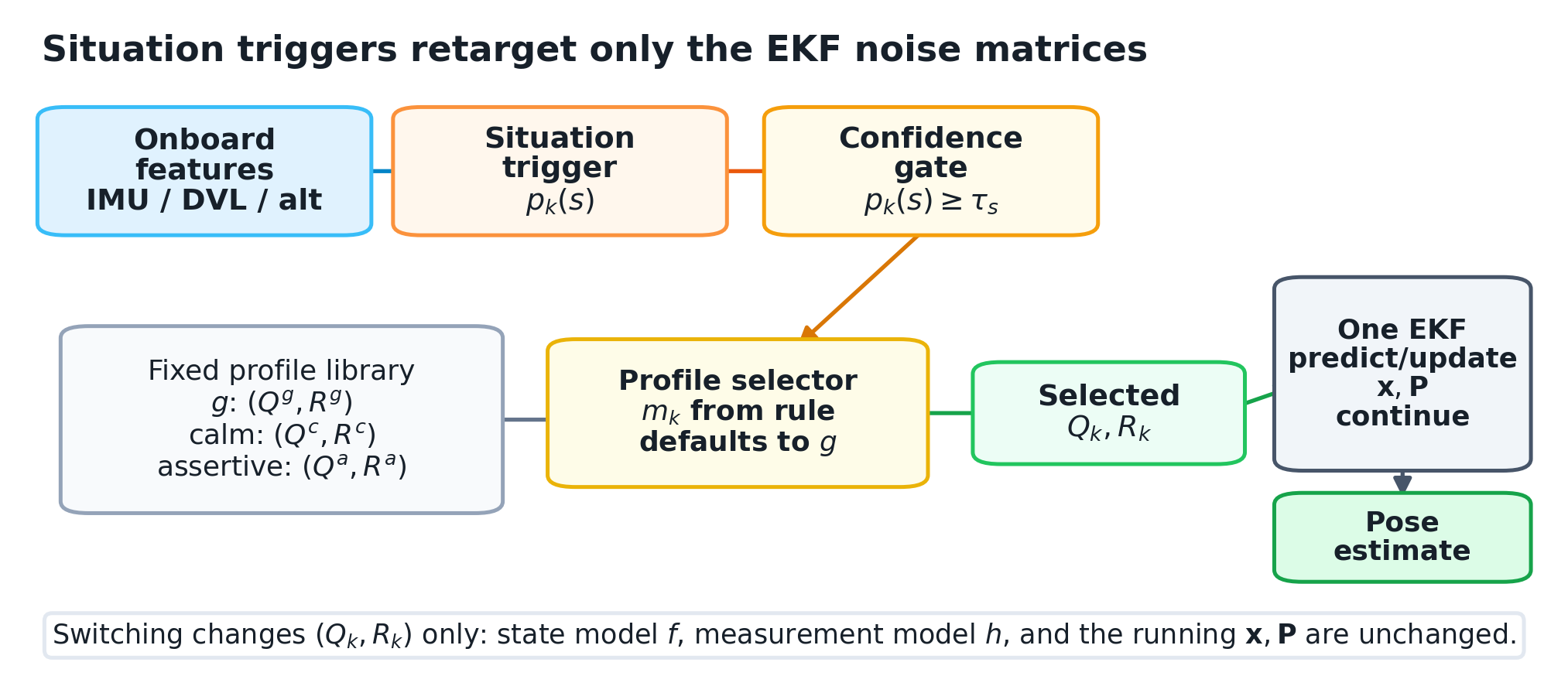}
\caption{Situation-gated $Q/R$ profile update. Onboard features produce situation probabilities; a validation-selected confidence gate either selects a fixed situation profile or keeps the global CAR-EKF profile. The selected $Q_k$ and $R_k$ enter the ordinary EKF covariance terms, while the state, covariance history, transition model, and measurement model remain continuous.}
\label{fig:qr-profile-update}
\end{figure}

At each step the trigger emits situation probabilities $p_k(s)$ over the situation set $\mathcal{S}$. Let $g$ denote the global CAR-EKF profile, and let $\rho(s)$ map an eligible situation to its fixed-matrix profile. ST-CAR-EKF uses the trigger only to choose the covariance profile,
\begin{equation}
\begin{aligned}
\hat{s}_k &= \arg\max_{s\in\mathcal{S}} p_k(s),\\
m_k &=
\begin{cases}
\rho(\hat{s}_k), & \hat{s}_k\in\mathcal{S}_{\mathrm{elig}},\; p_k(\hat{s}_k)\ge \tau_{\hat{s}_k},\\
g, & \text{otherwise},
\end{cases}\\
(Q_k,R_k) &= (Q^{m_k},R^{m_k}).
\end{aligned}
\label{eq:profile-select}
\end{equation}
The EKF itself is otherwise unchanged. The selected matrices enter only through the standard covariance terms,
\begin{equation}
\begin{aligned}
\hat{\mathbf{x}}_{k|k-1} &= f(\hat{\mathbf{x}}_{k-1|k-1},\mathbf{u}_k),\\
\mathbf{P}_{k|k-1} &= \mathbf{F}_k\mathbf{P}_{k-1|k-1}\mathbf{F}_k^\top
    + \mathbf{G}_k Q_k \mathbf{G}_k^\top,\\
\mathbf{S}_k &= \mathbf{H}_k\mathbf{P}_{k|k-1}\mathbf{H}_k^\top + R_k,
\end{aligned}
\label{eq:ekf-qr-injection}
\end{equation}
before the usual Kalman gain and correction are applied. A switch therefore changes the filter's uncertainty model, not its state model, measurement model, state vector, or covariance history. Through every switch the propagated state $\mathbf{x}$ and covariance $\mathbf{P}$ continue uninterrupted. Fig.~\ref{fig:qr-profile-update} illustrates this $Q/R$ injection point.

\subsection{Cross-Run Calibration Transfer}
\label{sec:transfer}
Noise calibration should be interpreted as a model fit on calibration data whose value is measured by cross-run transfer within the same pool benchmark, a small transfer-learning problem \cite{PanYang2010TransferLearning}. The DVL yaw correction and every $Q$/$R$ profile are estimated on the calibration (train) runs under sparse AprilTag supervision and then frozen; the validation split chooses among candidates; the test split is held-out pool data that measures calibration generalization to other runs from this dataset. This lens makes the comparison between validation and test results meaningful as a cross-run generalization gap rather than as two unrelated numbers, and it sets up the practical question of how little supervised data the calibration requires. In this dataset, 64 sparse velocity pairs suffice to estimate the DVL yaw correction and a two-run validation budget is enough to choose a policy that transfers to held-out pool runs. That is encouraging, but it is not yet a coverage theorem, cross-environment transfer claim, or open-water validation; we treat a controlled coverage-versus-accuracy ablation as the natural next study in Section~\ref{sec:discussion}.

\subsection{Selecting the Deployable Policy}
The routing table and the per-situation thresholds $\tau_s$ are selected on validation only, by a grid search over profile candidates followed by coordinate descent on the thresholds. The objective is the same label-weighted translation RMSE used for evaluation,
\begin{equation}
(\rho^\star,\boldsymbol{\tau}^\star)
= \arg\min_{\rho,\boldsymbol{\tau}}
\frac{\sum_{b\in\mathcal{B}_{\mathrm{val}}} n_b\,\mathrm{RMSE}_b(\rho,\boldsymbol{\tau})}
{\sum_{b\in\mathcal{B}_{\mathrm{val}}} n_b},
\label{eq:validation-policy}
\end{equation}
where $n_b$ is the number of matched labels in validation bag $b$ and $\mathrm{RMSE}_b$ is the run-level translation RMSE. The metric is therefore a weighted mean of per-run RMSE values, not a pooled row-level RMSE. The resulting deployable contract appears in Table~\ref{tab:stcar-policy}.

\begin{table}[!t]
\centering
\caption{Frozen deployable routing: per-situation fixed-matrix profile and validation confidence gates.}
\label{tab:stcar-policy}
\begin{tabular}{lcc}
\toprule
Situation & Fixed-matrix profile & Activation threshold \\
\midrule
straight line & calm process & 0.00 \\
transition mixed & assertive fusion & 0.35 \\
turning & assertive fusion & 0.99 (conservative) \\
\midrule
default & global & -- \\
\bottomrule
\end{tabular}
\end{table}

Straight-line maps to calm-process and is always engaged when predicted; transition/mixed maps to assertive-fusion behind the frozen deployment threshold $\tau=0.35$. Turning also maps to assertive-fusion because validation slices show that this profile is lower than global on turning segments (\SI{0.267}{\meter} versus \SI{0.272}{\meter}); the deployment question is how often that profile should activate. On held-out test the turning classifier is usable (F1 \num{0.79}) and mean $P(\text{turning})$ on predicted-turning rows reaches \num{0.86}, yet a one-at-a-time threshold sweep shows that opening the gate broadly helps validation (\SI{0.299}{\meter} versus \SI{0.318}{\meter} label-weighted per-run RMSE) but not test (best at the frozen $\tau=0.99$, \SI{0.471}{\meter} versus \SI{0.488}{\meter} global). The frozen policy therefore keeps a conservative $\tau=0.99$: assertive-fusion engages on 242 near-certain turning label rows across analyzed cohorts, while lower-confidence turning predictions fall back to global rather than risk the over-switching observed on test. Transition/mixed is the weaker class (held-out F1 \num{0.42}), but its $\tau=0.35$ gate is stable on both splits and, together with straight-line calm-process activation, carries the deployable gain.

\FloatBarrier
\section{Evaluation Protocol}
\subsection{Runtime Boundary}
\begin{figure}[!htbp]
\centering
\includegraphics[width=\textwidth]{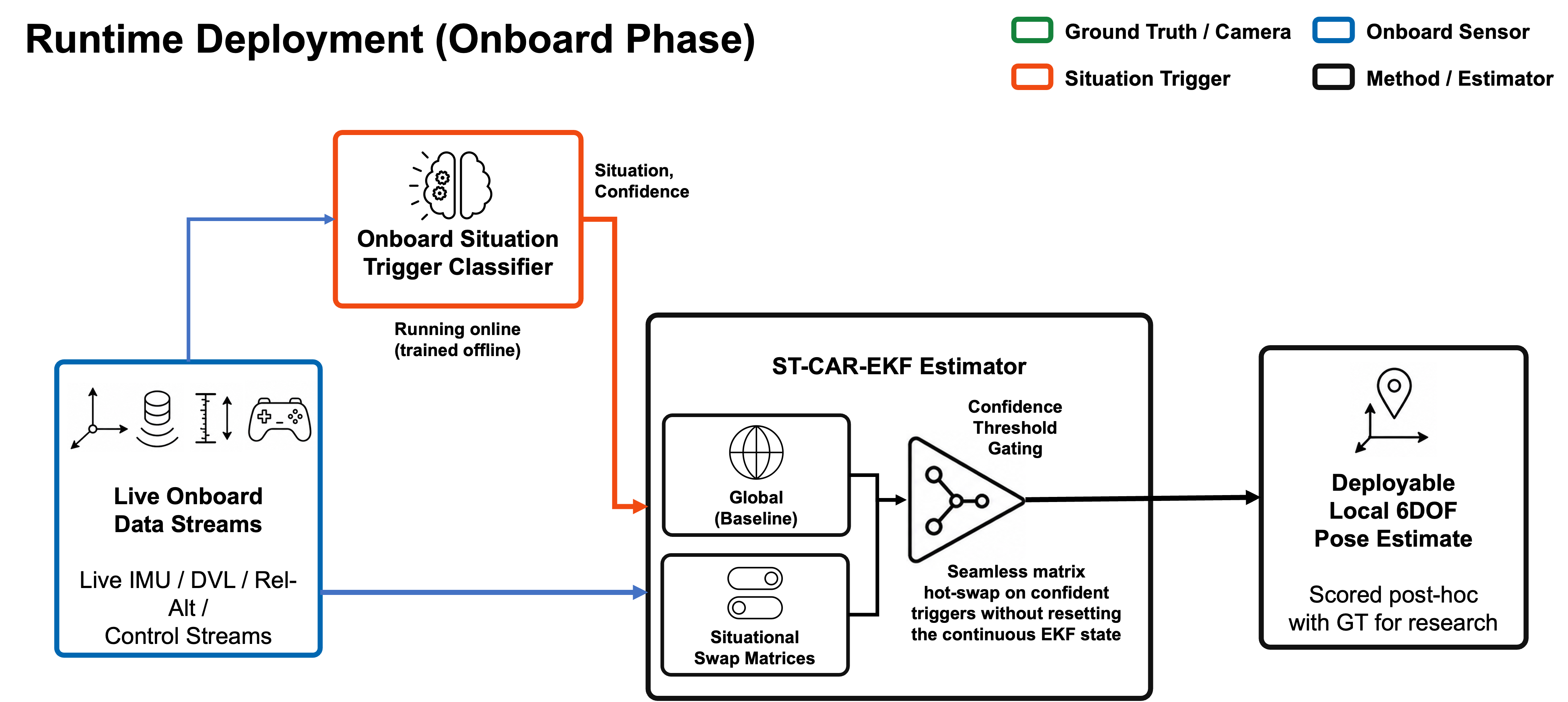}
\caption{Deployable ST-CAR-EKF runtime layout. Live onboard streams (IMU, DVL, relative altitude, control) feed both an onboard situation trigger classifier and one continuous error-state EKF. Confidence-gated triggers select situational swap matrices or the global baseline; matrix hot-swaps occur without resetting the continuous filter state. The pose estimate is vision-free at runtime and scored against AprilTag labels only post hoc.}
\label{fig:architecture}
\end{figure}

Every method is audited by what it may read at runtime. AprilTag detections, camera-derived poses, the tag map, ground-truth residuals, bag identity, and future samples are forbidden; the situation trigger is the only situation signal that survives that boundary (Fig.~\ref{fig:pipeline}, offline-only paths). Fig.~\ref{fig:architecture} shows the deployable runtime layout: live IMU, DVL, relative-altitude, and control streams feed an onboard situation trigger and one continuous error-state EKF. Because every switch happens inside one continuous filter, any reported gain is attributable to the noise-model swap and not to a convenient re-initialization.

\subsection{Metrics}
\label{sec:metrics}
We score absolute pose at sparse AprilTag label times after expressing each trajectory in the tag-map frame. The \emph{primary} headline is the \textbf{matched-label-count-weighted mean per-run translation RMSE}: for each run we compute translation RMSE (T-RMSE) over matched labels, then form a weighted mean across runs with weights proportional to each run's matched label count, so longer labeled segments count proportionally and short runs are not overweighted. Later references call this the \emph{label-weighted per-run translation RMSE}. Of the 1{,}178 curated test labels, 1{,}122 have a backward match to an EKF state estimate within \SI{0.2}{\second}; the remaining 56 labels occur before the first available filter state in two runs and are excluded consistently from the primary tables and bootstrap analyses.

We also report \textbf{yaw RMSE} and \textbf{rotation RMSE} (degrees, full attitude error) and vertical (Z) RMSE, because dead-reckoning quality is not only horizontal position. Absolute error at sparse tags measures global pose at observation times, but it is sensitive to the one-time map alignment and says little about how fast error accumulates locally. We therefore add three dead-reckoning-native measures that the literature relies on, all derived from a single relative-pose computation. First, \textbf{fixed-horizon relative-pose error} (RPE) in standard odometry-benchmark form~\cite{Geiger2012KITTI,Sturm2012TUM}: for every pair of labels separated by a fixed horizon (\SIlist{1;3;5;10}{\second}) we form the body-frame relative transform of both the estimate and ground truth and report the root-mean-square translation and rotation of their difference. Because RPE is computed from \emph{relative} transforms it is invariant to the global alignment, so it isolates local drift from any registration error. Second, \textbf{distance-normalized drift}: the same relative translation error expressed as a percentage of the ground-truth distance traveled over the horizon, which makes the number comparable across platforms and trajectory lengths. Third, the \textbf{error distribution} at sparse labels (median, \num{95}th percentile, and maximum translation error), because a navigation estimate is judged by both average performance and tail behavior.

The commercial stream corresponds to the recorded onboard DVL dead-reckoning topic documented by Water Linked \cite{WaterLinkedDVLDeadReckoning}: a Kalman-style fusion of DVL velocity and IMU/AHRS in a reset-relative frame. Its internals are unavailable, so it remains a black-box reference rather than a reproducible method. We report a \textbf{commercial-logged cohort} of three pool runs for in-house comparison, restrict commercial-stream scores to the two runs where it has usable predictions, and separately show individual runs; a single-run commercial-reference number must not be compared to a four-run in-house aggregate.

For uncertainty on the primary ST-CAR-EKF versus Global CAR-EKF comparison, we use paired bootstrap resampling \cite{Efron1979Bootstrap} of candidate-minus-baseline RMSE deltas on common matched rows. To reduce the effect of temporal correlation, the main interval resamples \SI{10}{\second} paired segments within each bag rather than individual rows; we also report a bag-level paired bootstrap. Both use 2000 bootstrap samples, fixed random seeds, and the paper's label-weighted per-run RMSE convention inside each resample. Negative deltas favor ST-CAR-EKF.

\FloatBarrier
\section{Results}
\subsection{Selecting the Global Baseline}
\begin{figure}[!htbp]
\centering
\includegraphics[width=\textwidth]{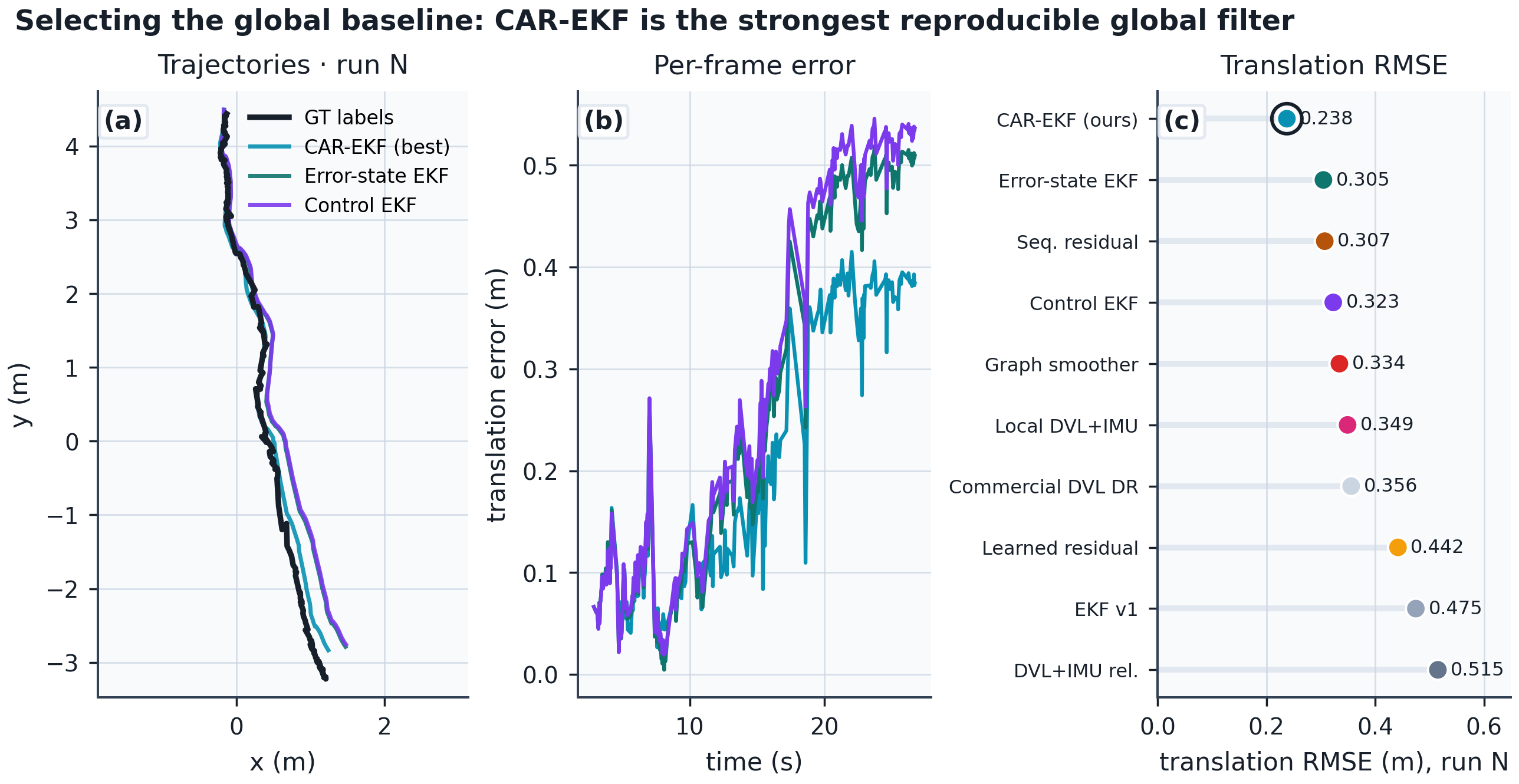}
\caption{Global baseline selection on the test run where the commercial stream is logged. (a) Estimated versus ground-truth trajectory and (b) per-frame translation error for the leading filters. (c) Same-run translation RMSE comparison: the calibrated EKF is the most accurate reproducible estimator on this run and is also the best global filter across the full held-out test split.}
\label{fig:global-benchmark}
\end{figure}

Among the reproducible vision-free filters, the globally calibrated EKF is the strongest single configuration on the held-out test split (\SI{0.488}{\meter} label-weighted per-run translation RMSE, ahead of the next filter at \SI{0.518}{\meter}). This Global CAR-EKF uses the same filter backbone as ST-CAR-EKF but keeps one validation-selected $Q/R$ profile for every timestep. Fig.~\ref{fig:global-benchmark} shows trajectories, per-frame error, and same-run RMSE on the test run where the commercial stream is logged: the calibrated EKF is the most accurate reproducible estimator on that run, while the commercial stream is included only as a black-box reference. We therefore adopt it as the global fallback that ST-CAR-EKF defaults to.

\subsection{Iterating the Switching Strategy}
Table~\ref{tab:stcar-iteration} compares switching strategies inside the same continuous filter. The simple strategies do not transfer reliably. Oracle routing, which picks each situation's profile from perfect ground-truth labels, improves validation but does \emph{not} beat the global baseline on test (\SI{0.494}{\meter} versus \SI{0.488}{\meter}); ungated triggering behaves the same way. Only when validation-selected per-situation confidence gates are added does a strategy transfer a gain to held-out test (validation \SI{0.318}{\meter} versus \SI{0.320}{\meter}; test \SI{0.471}{\meter} versus \SI{0.488}{\meter}); only that confidence-gated strategy beats the global baseline on both splits. This result indicates that switching helps only when the filter requires sufficient trigger confidence before changing profiles.

\begin{table}[!t]
\centering
\caption{Switching-strategy comparison within one continuous EKF. Values are label-weighted per-run translation RMSE in meters.}
\label{tab:stcar-iteration}
\footnotesize
\begin{tabular}{@{}>{\raggedright\arraybackslash}p{0.46\linewidth}>{\centering\arraybackslash}p{0.15\linewidth}>{\centering\arraybackslash}p{0.12\linewidth}>{\centering\arraybackslash}p{0.12\linewidth}@{}}
\toprule
Strategy & Runtime signal & Val. RMSE & Test RMSE \\
\midrule
Global CAR-EKF (single fixed profile) & no & 0.320 & 0.488 \\
Fixed-matrix swap, oracle GT (upper bound) & GT labels & 0.308 & 0.494 \\
Fixed-matrix swap, trigger, no gate & trigger & 0.301 & 0.494 \\
\textbf{Fixed-matrix swap, trigger + confidence (ours)} & trigger & 0.318 & 0.471 \\
\bottomrule
\end{tabular}
\end{table}

\subsection{Situation Profiles on Validation}
Before deploying triggers, we ask whether distinct fixed-matrix profiles help on their motion regimes. Table~\ref{tab:situation-validation} compares validation-only, situation-conditioned translation RMSE on ground-truth-labeled segments of the two validation runs against the global profile on the same segments. Assertive-fusion is lower than global on turning and mixed-transition slices; straight-line segments favor the global profile on this split. The deployable policy therefore activates calm-process on straight-line triggers and assertive-fusion on mixed transitions behind the frozen $\tau=0.35$ threshold, and it \emph{reserves} the same assertive-fusion profile for turning behind a conservative $\tau=0.99$ until broader activation is shown to transfer to held-out pool runs.

\begin{table}[!t]
\centering
\caption{Validation-only situation-conditioned translation RMSE (m) on GT-labeled validation segments, compared with the global profile on the same segments.}
\label{tab:situation-validation}
\footnotesize
\begin{tabular}{@{}>{\raggedright\arraybackslash}p{0.28\linewidth}>{\raggedright\arraybackslash}p{0.22\linewidth}>{\centering\arraybackslash}p{0.16\linewidth}>{\centering\arraybackslash}p{0.16\linewidth}@{}}
\toprule
Situation & Selected profile & Situation slice & Global on slice \\
\midrule
turning & assertive fusion & \textbf{0.267} & 0.272 \\
transition mixed & assertive fusion & \textbf{0.268} & 0.273 \\
straight line & global & 0.262 & 0.262 \\
\bottomrule
\end{tabular}
\end{table}

\subsection{Confidence Gating}
\begin{figure}[!htbp]
\centering
\includegraphics[width=0.95\textwidth]{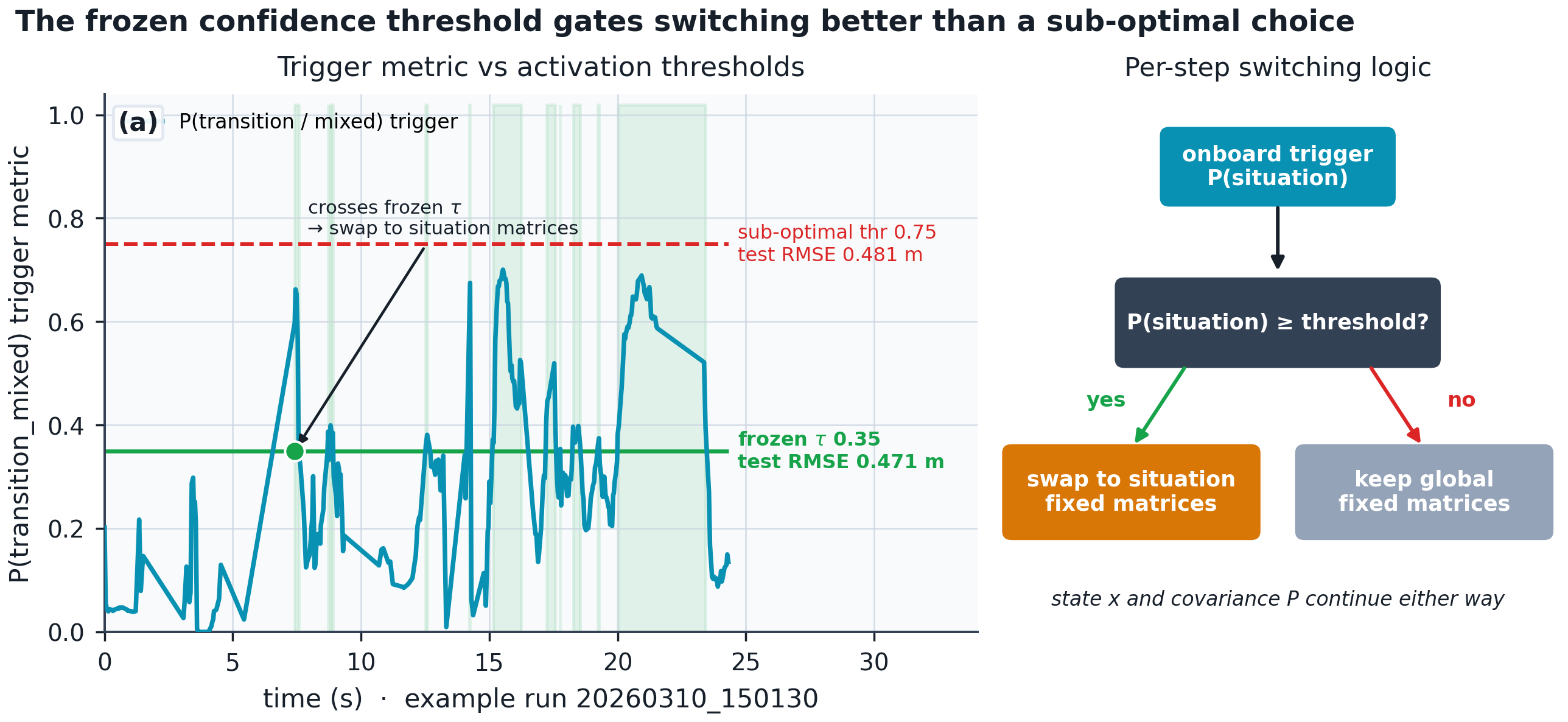}
\caption{Confidence-gated switching. (a) The onboard transition/mixed trigger probability over time on an example run; a swap fires when it crosses the frozen deployment threshold $\tau=0.35$ (shaded region = deployed policy active). A sub-optimal threshold yields higher held-out test RMSE than the frozen validation-selected threshold. (b) Per-step logic: when the probability clears its threshold the EKF swaps in the situation's fixed matrices, otherwise it keeps the global matrices; state and covariance are never reset.}
\label{fig:confidence}
\end{figure}

Fig.~\ref{fig:confidence} illustrates the confidence gate. The onboard transition/mixed trigger probability is plotted over time on an example run, and a profile swap fires whenever the metric crosses the frozen deployment threshold $\tau=0.35$. A poorly chosen threshold switches at unsupported times and increases held-out RMSE, while the validation-selected threshold is the deployed policy. A one-at-a-time diagnostic sweep finds a slightly lower test-only transition/mixed threshold at $\tau=0.45$ (\SI{0.470}{\meter} versus \SI{0.471}{\meter}), but that value is not used because the policy is frozen from validation before held-out scoring. Either way the continuous state and covariance carry across the decision, so the gate tunes \emph{when} to change the noise model rather than whether to disturb the estimate.

\subsection{Held-Out Test Benchmark}
ST-CAR-EKF is the best deployable in-house estimator on the held-out test split (Table~\ref{tab:stcar-test}), improving label-weighted per-run translation RMSE over the same CAR-EKF backbone with one global profile (\SI{0.471}{\meter} versus \SI{0.488}{\meter}) with yaw and rotation essentially unchanged. Scores for the commercial stream are reported separately in Table~\ref{tab:vendor-cohort} and must not be compared to this four-run in-house aggregate. It wins all four held-out runs by per-run T-RMSE; per-run held-out results are provided in Table~S1 of the supplementary material. A paired bootstrap gives the same modest but nonzero direction at two resolutions: \SI{-0.017}{\meter} candidate-minus-baseline delta with a 95\% confidence interval of $[-0.024,-0.008]~\si{\meter}$ on \SI{10}{\second} segments and $[-0.026,-0.006]~\si{\meter}$ when resampling bags. On the held-out split, the same ordering holds under the alignment-independent relative-pose metrics reported below, so the gain is consistent with reduced local translational drift rather than a favorable map registration. Oracle routing from perfect ground-truth labels does not beat the global baseline on test (Table~\ref{tab:stcar-iteration}), which is why the paper stresses confidence-gated onboard switching rather than switching in the abstract.

\begin{table}[!t]
\centering
\begin{threeparttable}
\caption{Held-out test pose error over four runs (1{,}122 matched label rows). Label-weighted per-run aggregates.}
\label{tab:stcar-test}
\footnotesize
\begin{tabular}{@{}>{\raggedright\arraybackslash}p{0.36\linewidth}>{\centering\arraybackslash}p{0.12\linewidth}>{\centering\arraybackslash}p{0.11\linewidth}>{\centering\arraybackslash}p{0.11\linewidth}>{\centering\arraybackslash}p{0.11\linewidth}@{}}
\toprule
Method & T-RMSE (m) & Z-RMSE (m) & Yaw (deg) & Rot.\ (deg) \\
\midrule
\textbf{ST-CAR-EKF (ours)$^{\star}$} & 0.471 & 0.058 & 2.73 & 4.01 \\
Global CAR-EKF & 0.488 & 0.059 & 2.73 & 4.01 \\
Oracle-GT fixed-matrix route & 0.494 & 0.058 & 2.73 & 4.01 \\
Offline error-state EKF & 0.518 & 0.059 & 2.72 & 4.01 \\
Offline control EKF & 0.531 & 0.072 & 2.72 & 4.00 \\
Local DVL+IMU odometry & 0.625 & 0.124 & 2.67 & 3.99 \\
\bottomrule
\end{tabular}
\begin{tablenotes}[flushleft]
\footnotesize
\item[$^{\star}$] Proposed method. Bold indicates the best deployable in-house estimator in each column.
\end{tablenotes}
\end{threeparttable}
\end{table}

\subsection{Commercial DVL Dead Reckoning: Accuracy and Robustness}
\begin{figure}[!htbp]
\centering
\includegraphics[width=\textwidth]{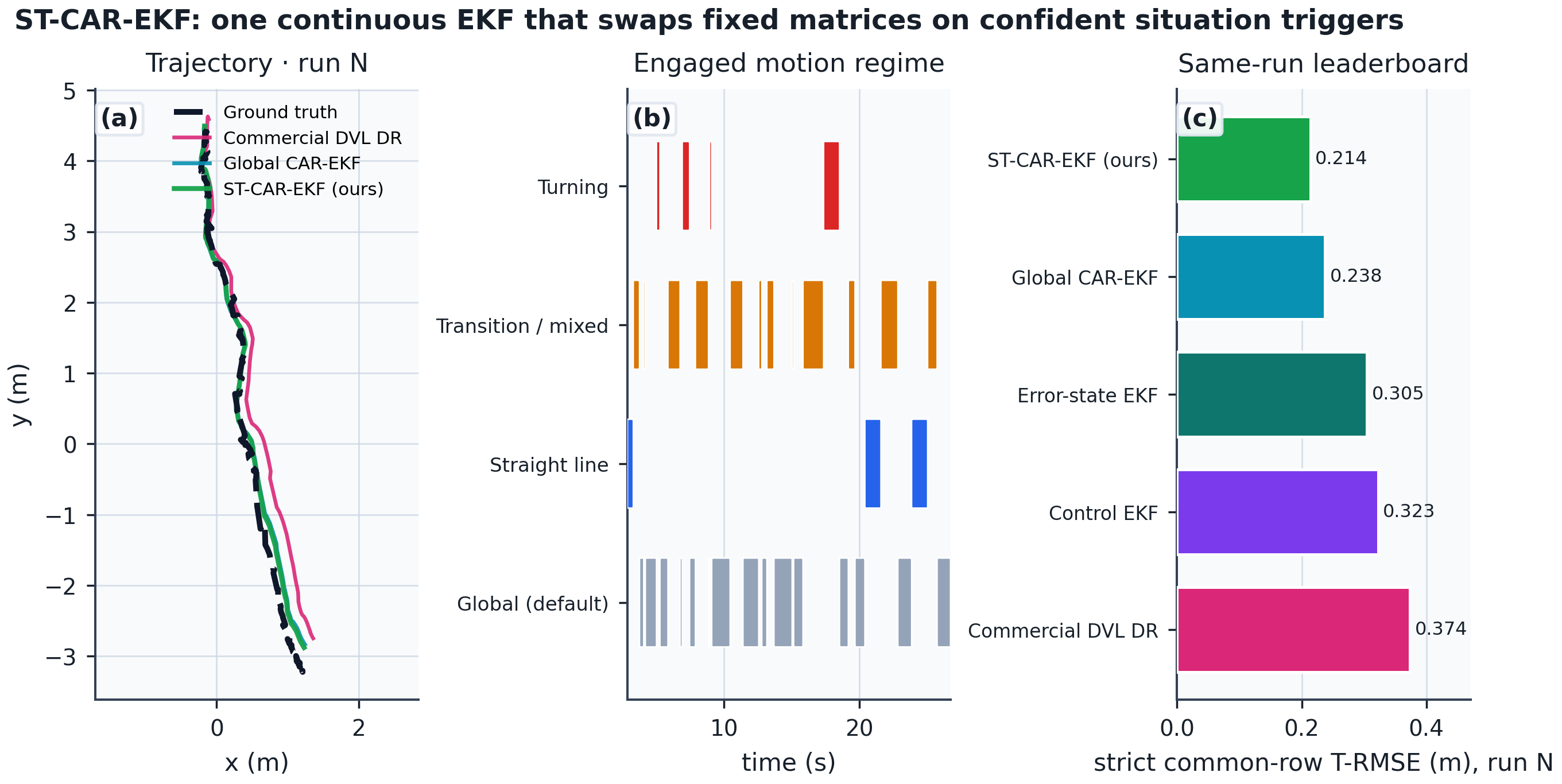}
\caption{Equal-footing comparison on test run N, where the commercial stream is logged. (a) ST-CAR-EKF (green) tracks the ground-truth path while Commercial DVL DR drifts. (b) Engaged motion regime over time: global by default, situation-specific fixed matrices only on confident triggers. (c) Translation RMSE on strict common label rows: ST-CAR-EKF (\SI{0.214}{\meter}) is lower than the global CAR-EKF (\SI{0.238}{\meter}) and Commercial DVL DR (\SI{0.374}{\meter}) on this run. Across the full four-run held-out test split ST-CAR-EKF remains the best deployable in-house estimator (\SI{0.471}{\meter} versus \SI{0.488}{\meter}).}
\label{fig:final}
\end{figure}

Table~\ref{tab:vendor-cohort} summarizes the commercial-logged cohort. The in-house cohort contains three pool runs, while commercial-stream scores are restricted to the two runs where usable predictions exist. On the three-run in-house cohort, ST-CAR-EKF achieves the lowest label-weighted per-run translation RMSE (\SI{0.257}{\meter} versus \SI{0.274}{\meter} for the global CAR-EKF on the same 1{,}061 matched rows). As a black-box reference, the commercial stream exposes failure modes but is not a reproducible baseline: on validation run J it exceeds \SI{1.29}{\meter} translation RMSE while ST-CAR-EKF stays near \SI{0.11}{\meter}, and cohort yaw error is \SI{24.8}{\degree} for the commercial stream versus \SI{2.6}{\degree} for ST-CAR-EKF. On the strict common-row comparison over the two runs with commercial predictions (451 labels), ST-CAR-EKF (\SI{0.167}{\meter}) is lower than global (\SI{0.183}{\meter}) and the commercial stream (\SI{0.786}{\meter}). Relative to the nominal BlueROV2 body length, these errors correspond to approximately 0.37 body lengths for ST-CAR-EKF, 0.40 body lengths for the global CAR-EKF, and 1.72 body lengths for the commercial stream. On test run N, an equal-footing comparison on strict common label rows places ST-CAR-EKF at \SI{0.214}{\meter}, the global filter at \SI{0.238}{\meter}, and the commercial stream at \SI{0.374}{\meter} (Fig.~\ref{fig:final}).

\begin{table}[!t]
\centering
\begin{threeparttable}
\caption{Commercial-logged cohort pose error (three in-house runs, 1{,}061 matched rows). Label-weighted per-run aggregates.}
\label{tab:vendor-cohort}
\footnotesize
\begin{tabular}{@{}>{\raggedright\arraybackslash}p{0.30\linewidth}>{\centering\arraybackslash}p{0.13\linewidth}>{\centering\arraybackslash}p{0.11\linewidth}>{\centering\arraybackslash}p{0.11\linewidth}>{\centering\arraybackslash}p{0.11\linewidth}@{}}
\toprule
Method & T-RMSE (m) & Z-RMSE (m) & Yaw (deg) & Rot.\ (deg) \\
\midrule
\textbf{ST-CAR-EKF (ours)} & 0.257 & 0.048 & 2.55 & 3.96 \\
Global CAR-EKF & 0.274 & 0.049 & 2.55 & 3.95 \\
Commercial DVL DR$^{\dagger}$ & 0.753 & 0.190 & 24.83 & 26.51 \\
\bottomrule
\end{tabular}
\begin{tablenotes}[flushleft]
\footnotesize
\item[$\dagger$] Commercial DVL DR denotes the logged commercial stream. Its internals are unavailable, so it is treated as a black-box reference.
\end{tablenotes}
\end{threeparttable}
\end{table}

\subsection{Relative Drift and Error Distribution}
Absolute pose error can flatter or punish a method through the global map alignment, so we check the held-out-test finding with the alignment-independent measures of Table~\ref{tab:relative-drift}. On the held-out test split, ST-CAR-EKF posts the lowest relative-pose translation error at every horizon (\SI{0.166}{\meter} at \SI{5}{\second} versus \SI{0.172}{\meter} for the global filter and \SI{0.189}{\meter}/\SI{0.187}{\meter} for the offline error-state and control variants), which corresponds to a local drift of roughly \SI{9}{\percent} of distance traveled; relative rotation drift is essentially level across the in-house filters ($\approx\SI{2.6}{\degree}$). On the commercial-logged cohort, Global CAR-EKF is lower on some in-house relative-translation metrics, while the commercial stream separates most clearly by relative rotation drift and tail error.

The relative view is most revealing on the commercial-logged cohort as a robustness diagnostic rather than a universal in-house win count. ST-CAR-EKF has the best absolute translation RMSE on that cohort (\SI{0.257}{\meter} versus \SI{0.274}{\meter} for the global CAR-EKF), but the global filter has lower \SI{5}{\second} relative translation drift in Table~\ref{tab:relative-drift} (\SI{0.139}{\meter} versus \SI{0.168}{\meter}). The commercial stream, ST-CAR-EKF, and the in-house filters share a comparable \emph{median} absolute error ($\approx\SI{0.22}{\meter}$), so on a typical label the black-box reference looks competitive. The separation lives in the tail and in orientation: ST-CAR-EKF holds \SI{5}{\second} relative rotation drift to \SI{2.95}{\degree} and a worst-case absolute error of \SI{0.58}{\meter}, whereas the commercial stream drifts \SI{31.94}{\degree} in relative rotation and reaches \SI{3.57}{\meter} at its worst, with a \num{95}th percentile of \SI{2.65}{\meter} against our \SI{0.51}{\meter}. Because these are relative quantities, the commercial degradation cannot be explained away as a frame-registration artifact on these runs. This is the kind of black-box reference robustness failure that motivates reporting common-row commercial-stream results separately from the reproducible four-run test benchmark.

\begin{table}[!t]
\centering
\begin{threeparttable}
\caption{Dead-reckoning-native evaluation at the \SI{5}{\second} horizon and absolute translation-error tails. Bold indicates the best value within each cohort and column.}
\label{tab:relative-drift}
\footnotesize
\begin{tabular}{@{}>{\raggedright\arraybackslash}p{0.30\linewidth}ccccc@{}}
\toprule
Method & Rel.\ T-RMSE (m) & Drift (\% dist.) & Rel.\ Rot (deg) & p95 (m) & max (m) \\
 & @\SI{5}{\second} & @\SI{5}{\second} & @\SI{5}{\second} & abs. & abs. \\
\midrule
\multicolumn{6}{@{}l}{Held-out test split (4 runs, 1{,}122 matched rows)}\\
ST-CAR-EKF (ours)$^{\star}$ & \textbf{0.166} & \textbf{9.0} & 2.65 & 1.17 & \textbf{1.26} \\
Global CAR-EKF & 0.172 & 9.3 & 2.63 & 1.18 & 1.29 \\
Offline error-state EKF & 0.189 & 10.4 & 2.64 & \textbf{1.14} & \textbf{1.26} \\
Offline control EKF & 0.187 & 10.2 & \textbf{2.60} & 1.15 & 1.27 \\
\midrule
\multicolumn{6}{@{}l}{Commercial-logged cohort (3 runs, $\approx$1{,}061 matched rows)}\\
ST-CAR-EKF (ours)$^{\star}$ & 0.168 & 17.9 & 2.95 & 0.51 & 0.58 \\
Global CAR-EKF & \textbf{0.139} & \textbf{10.5} & 2.81 & \textbf{0.39} & \textbf{0.41} \\
Offline error-state EKF & 0.157 & 11.7 & 2.81 & 0.50 & 0.52 \\
Offline control EKF & 0.153 & 11.4 & \textbf{2.73} & 0.53 & 0.55 \\
Commercial DVL DR$^{\dagger}$ & 0.551 & 19.5 & 31.94 & 2.65 & 3.57 \\
\bottomrule
\end{tabular}
\begin{tablenotes}[flushleft]
\footnotesize
\item[$^{\star}$] Proposed method.
\item[$\dagger$] Commercial DVL DR denotes the logged commercial stream; it is a black-box reference with unavailable internals.
\item Cohort headers identify held-out test (four runs) versus commercial-logged (three runs) evaluation sets.
\end{tablenotes}
\end{threeparttable}
\end{table}

\FloatBarrier
\section{Discussion}
\label{sec:discussion}
\subsection{What the Results Support}
Three claims hold up, and they answer different questions a reviewer might ask. On \textbf{accuracy}, retargeting only the fixed noise matrices of a strong continuous filter, with no state reset, yields a modest but consistent held-out test gain over the same backbone with a single global profile. On \textbf{black-box reference robustness}, the commercial stream has larger rotation drift and tail errors on available common rows, but that comparison is deliberately separated from the reproducible in-house benchmark. The gap is not only a registration artifact, because the alignment-independent relative-pose metrics show much larger commercial-stream relative rotation drift and worst-case error (Table~\ref{tab:relative-drift}). On \textbf{situation headroom}, validation shows that situational fixed matrices can be lower than the global matrix on turning and mixed-transition segments (Table~\ref{tab:situation-validation}), and confidence-gated switching is what makes that structure survive held-out evaluation in this dataset.

\subsection{Why Situation Separation Matters}
Blind switching is not enough. Oracle routing and ungated triggering both underperform the global baseline on held-out test (Table~\ref{tab:stcar-iteration}), which means the value is not ``swap more often'' but swap only when an onboard trigger is confident. The validation slice comparison (Table~\ref{tab:situation-validation}) is the structural evidence: some motion slices prefer different fixed noise profiles, and confidence-gated switching is the mechanism that turns that validation structure into a held-out gain.

The activation thresholds are themselves a quantity we are still learning to set, and they are limited as much by transfer evidence as by the trigger itself. Turning is the clearest example: validation shows a profile advantage on turning segments, and the classifier is already usable on test (F1 \num{0.79}), yet opening the gate aggressively fails to transfer (Table~\ref{tab:stcar-policy}). A high $\tau=0.99$ is therefore not a claim that turning is irrelevant; it is a deployability choice to fire assertive-fusion only when $P(\text{turning})$ is near-certain until more validation coverage or calibration supports a lower gate without the test regression seen when turning is ungated. As trigger probabilities become better calibrated, the same slice-level accuracy should be reachable at a lower threshold, which would recover more of the per-situation headroom without changing the runtime contract.

\subsection{Which Covariance Choices Actually Matter}
A reasonable reviewer will ask which parts of the noise model are worth all this effort, and our calibration already gives a qualitative answer. The two profiles that survived selection move along exactly two axes: the process-noise level that governs how much the filter trusts its own propagation on straight legs, and the DVL measurement trust that governs transitions and turns. Just as telling is what did \emph{not} survive. The Huber robust kernel and the innovation-adaptive noise term were available throughout the search and were not selected on this data, which suggests that on calm pool runs the dominant levers are the static $Q$ and $R$ magnitudes rather than outlier machinery. We present this as evidence, not proof. A formal sensitivity sweep, perturbing individual $Q$ and $R$ entries and tracking both RMSE and filter-consistency degradation, would let a designer rank which terms must be nailed and which are forgiving, and it is a clean, self-contained experiment we have deliberately scoped for the follow-on study rather than approximated here.

\subsection{How Much Ground Truth Is Enough}
The cross-run calibration-transfer framing (Section~\ref{sec:transfer}) puts a practical variable in the foreground: tag coverage. The present results are compatible with a frugal calibration story, since the DVL frame is fixed from 64 sparse velocity pairs and the noise model is chosen on two validation runs, yet the calibration still transfers to held-out pool runs from the same benchmark. They do not establish the minimum supervision required. The natural way to make this rigorous is a coverage-versus-accuracy curve that degrades the supervised label density used for calibration and re-measures held-out RMSE, which would turn ``partial fiducial coverage'' from a caveat into a deployment guideline for someone seeding a new site with only a few tags. Our dataset supports the bag-level version of this analysis directly, and a controlled subsampling ablation is the single most useful addition we can make next.

\subsection{Temporally Correlated DVL Noise}
The EKF treats DVL velocity error as white, which is the standard but imperfect assumption. DVL residuals are frequently correlated in time, especially in water-track mode or over textured near-bottom features, and a white-noise model cannot represent that structure between updates. Underwater model-aiding and correlation-aware INS/DVL work treat related structure explicitly \cite{ArnoldMedagoda2018ModelAided,Cohen2025CrossCorrelation}. We state this as a modeling assumption rather than claim to have solved it. A validation-only residual audit over the two validation bags logs per-update residuals and normalized innovation squared (NIS) for each measurement update. For DVL velocity, the measurement dimension is three, so the audit reports an NIS ratio $\mathrm{NIS}/3$ whose expected mean is approximately one under the assumed Gaussian covariance model. The global profile's DVL-velocity updates have mean NIS ratio \num{0.38} across 613 updates, which may indicate conservative covariance scaling, temporal residual correlation, model mismatch, or a combination. The same audit computes lag-1 autocorrelation for DVL residual components $x$, $y$, $z$, residual norm, and NIS ratio, then takes the maximum absolute component/run/profile value. The largest DVL residual component value is \num{0.92} on the $z$ residual for a calm-process validation segment, with mean absolute lag-1 DVL residual autocorrelation \num{0.59} across audited DVL entries. These diagnostics do not prove a single failure mechanism, but they support treating temporally correlated DVL error as a remaining modeling limitation. A principled remedy is to augment the process model with a first-order Gauss--Markov velocity-error state so that the filter carries the correlation explicitly. Quantifying the residual autocorrelation and weighing a Gauss--Markov augmentation against its added state is a concrete next contribution; for this paper we bound the issue and point to it rather than absorb it into the headline method.

\subsection{Why the Commercial Stream Stays a Reference}
The commercial stream represents a real onboard product path \cite{WaterLinkedDVLDeadReckoning}, but its internals are unavailable, so it cannot be a building block for a reproducible method. We score it as a black-box reference on the commercial-logged cohort. Under that protocol, ST-CAR-EKF is lower-error on the common-row comparison and less affected by the stream's failure run; a single-run test number can obscure that behavior, which is why we report the three-run cohort and per-run failures explicitly.

\subsection{Limitations}
The ground truth is sparse and visually gated, and the tag map is internally estimated rather than externally surveyed; relative-pose metrics are included to reduce, not eliminate, dependence on that map alignment. The data are collected in a pool, so currents, turbidity, biofouling, and open-water acoustic effects are absent; the claim is therefore a clean pool benchmark and not open-water validation. Turning is routed to a validated assertive-fusion profile but deployed behind a conservative $\tau=0.99$ because broader activation improves validation yet does not transfer to held-out pool runs; we do not retune that threshold from the test split. Finally, the commercial-reference comparison rests on three logged runs and only two strict common-row runs where the commercial stream has usable predictions, which makes it useful as a black-box reference but not a product-level verdict. These limits matter, but the core runtime boundary remains vision-free, the train/validation/test split is bag-level, and the main ST-CAR-EKF versus Global CAR-EKF comparison is reproducible on the four held-out runs.

\section{Conclusion}
We presented ST-CAR-EKF, a deployable, vision-free estimator that runs one continuous calibrated EKF and swaps only its fixed $Q$/$R$ matrices when an onboard situation trigger is confident, never resetting the filter. We framed the noise calibration as cross-run calibration transfer from sparse AprilTag supervision to held-out pool runs. On a reproducible BlueROV2 benchmark, evaluated with both absolute pose error and dead-reckoning-native relative-pose drift, the method gives a small but statistically supported held-out improvement over the same CAR-EKF backbone with one global profile. A restricted common-row comparison also shows substantially lower error than the logged commercial stream, which we treat as a black-box reference rather than a reproducible baseline. The clearest next steps are a coverage-versus-accuracy study for how little supervision calibration needs, a formal $Q$/$R$ sensitivity and consistency sweep, richer per-situation training, and an explicit treatment of temporally correlated DVL error.

\FloatBarrier
\section*{Acknowledgment}
This work was supported by the U.S. Army Corps of Engineers, Engineer Research and Development Center, Construction Engineering Research Laboratory, under Contract Award No.~W9132T249C011. The authors thank the members of the Autonomous and Unmanned Vehicle Systems Laboratory and the Center for Autonomous Construction and Manufacturing at Scale at the University of Illinois Urbana-Champaign for their support. The authors also acknowledge the U.S. Army Engineer Research and Development Center, Construction Engineering Research Laboratory, for technical collaboration and review of the manuscript. OpenAI Codex was used to assist with manuscript editing, formatting, and consistency checks; the authors reviewed the resulting changes and take responsibility for the final content.

\bibliographystyle{IEEEtran}
\bibliography{references}

\end{document}